\documentclass{article}
\usepackage{amsmath,graphicx,mlspconf}

\copyrightnotice{978-1-7281-0824-7/19/\$31.00 {\copyright}2019 IEEE}

\usepackage{amssymb,amsfonts,amsmath}
\usepackage{mathtools}
\usepackage[linesnumbered,ruled,vlined]{algorithm2e}
\usepackage{arydshln}
\usepackage{booktabs}
\usepackage[bookmarks=false]{hyperref}
\usepackage{tikz}
\usetikzlibrary{fit,calc,positioning,decorations.pathreplacing,matrix}
\usetikzlibrary{shapes,arrows}
\usetikzlibrary{calc}
\usepackage{multirow}
\usepackage{stackengine}

\hypersetup{
    colorlinks,
    linkcolor={black},
    citecolor={black},
    urlcolor={black},
}

\usepackage{xcolor}

\usepackage{enumitem}

\newcommand{\w}{\mathbf{w}}
\newcommand{\smlv}{\mathbf{v}}

\newcommand{\smla}{\mathbf{a}}
\newcommand{\smlu}{\mathbf{u}}
\newcommand{\smlq}{\mathbf{q}}
\newcommand{\x}{\mathbf{x}}
\newcommand{\z}{\mathbf{z}}
\newcommand{\y}{\mathbf{y}}

\newcommand{\rr}{\mathbf{r}}
\newcommand{\proj}{\mathbf{\Phi}}

\newcommand{\A}{\mathbf{A}}
\newcommand{\C}{\mathbf{C}}

\newcommand{\dee}{\mathbf{d}}

\newcommand{\R}{{\mathbb{R}}}

\newcommand{\cvec}{\mathbf{c}}
\newcommand{\eye}{\mathbf{I}}

\newcommand{\half}{{\textstyle \frac{1}{2}}}

\newcommand{\ip}[2]{\langle #1,#2 \rangle}

\title{RandNet: Deep Learning with Compressed Measurements of Images}
\name{Thomas Chang$^{*}$\hskip 1in Bahareh Tolooshams$^{*}$\hskip 1in Demba Ba\thanks{* The authors contributed equally to this work.}
}
\address{School of Engineering and Applied Sciences, Harvard University, Cambridge, MA}
\begin{document}
%

\maketitle
\begin{abstract}
Principal component analysis, dictionary learning, and auto-encoders are all unsupervised methods for learning representations from a large amount of training data. In all these methods, the higher the dimensions of the input data, the longer it takes to learn. 
We introduce a class of neural networks, termed RandNet, for learning representations using compressed random measurements of data of interest, such as images. RandNet extends the convolutional recurrent sparse auto-encoder architecture to dense networks and, more importantly, to the case when the input data are compressed random measurements of the original data. 
Compressing the input data makes it possible to fit a larger number of batches in memory during training. Moreover, in the case of sparse measurements, 
training is more efficient computationally. We demonstrate that, in unsupervised settings, RandNet performs dictionary learning using compressed data. In supervised settings, we show that RandNet can classify MNIST images with minimal loss in accuracy, despite being trained with random projections of the images that result in a $50\%$ reduction in size. Overall, our results provide a general principled framework for training neural networks using compressed data.
\end{abstract}
\section{Introduction}\label{sec:intro}
\vspace{-3mm}
Representation learning has become an important problem in recent years both in the signal processing and machine learning communities. In signal processing, dictionary learning (DL)~\cite{Agarwal2016LearningSU} is the de facto method for learning adaptive data representations. In machine learning, deep learning 
is the method of choice to learn representations that are adapted to data. In several tasks such as 
image classification
, the early success of DL, and more recently deep learning, can be attributed largely to their ability to learn representations tailored to the task of interest.

Traditionally, DL has been restricted to the shallow case, where only one set of weights is learned from data. The works in~\cite{TolooshamsBahareh2018SCDL,TolooshamsBahareh2019deepresidualAE} have shown a one-to-one correspondence between DL and neural networks (NNs) and, more specifically, how to train an auto-encoder to perform unsupervised DL. Recent work~\cite{PapyanV2017CNNA,BaDemba2018DSrt} has shown the connections between deep DL and deep learning, emphasizing a new perspective on deep NNs as efficient algorithms for solving deep DL problems. The primary advantages of using deep NNs for DL are the widespread GPU-based infrastructure that supports them (e.g.~TensorFlow), and the ease with which they can be deployed. However, in a number of applications of both DL and NNs, such as video processing, the size of datasets required for learning is a computational bottleneck. 

This motivates the need for a framework that can learn from reduced-size data. In~\cite{pmlr-v32-anaraki14,Pourkamali-AnarakiFarhad2015Edlv}, the authors propose two different methods for learning representations from random projections of data. The first,~\cite{pmlr-v32-anaraki14}, performs PCA using compressed measurements. The second, called CK-SVD~\cite{Pourkamali-AnarakiFarhad2015Edlv}, is an optimization-based alternating-minimization algorithm~\cite{Agarwal2016LearningSU} for DL from sparse random measurements of data. The benefits of these approaches are memory and computational efficiency~\cite{pmlr-v32-anaraki14}, and the ability to train using reduced-size data~\cite{Pourkamali-AnarakiFarhad2015Edlv}.

This paper brings together the idea of training NNs for DL, and that of using compressed random measurements of data to solve DL tasks. Specifically, we introduce a framework to train NNs to perform unsupervised DL or a supervised task, e.g.~classification, using only compressed projections of the data. The resulting class of architectures, which we call RandNet, is more efficient in terms of memory and computation than conventional ones. 
In the unsupervised setting, we demonstrate that RandNet performs DL. The architecture is a variant of the constrained recurrent sparse auto-encoder (CRsAE)~\cite{TolooshamsBahareh2019deepresidualAE} in which the encoder uses randomly projected data to obtain a sparse representation. 
For supervised tasks, RandNet uses the sparse representation produced at the output of the encoder as features. 
We demonstrate that the performance of RandNet in the classification of MNIST handwritten digits~\cite{mnist} rivals the state-of-the-art.

We begin the remainder of our treatment in the next section, where we introduce the DL problem. We introduce RandNet in Sec.~\ref{sec:randnets}, and apply it to DL and classification of MNIST digits in Sec.~\ref{sec:expts}. We conclude in Sec.~\ref{sec:conc}.


\vspace{-3mm}
\section{Dictionary Learning}\label{sec:dl}
\vspace{-3mm}
%
Let $\x^j \in \R^{p}$ be a sparse vector and $\y^j \in \R^{N}$ be the vector obtained as the sum of the sparse linear combination of columns of a dictionary $\A \in \R^{N \times p}$ and additive noise 
\vspace{-3mm}
\begin{equation} \label{eq:gen}
\y^j = \A \x^j + \smlv^j, j = 1,\cdots, J,
\vspace{-3mm}
\end{equation}

\begin{figure*}[htb]
	\vspace*{-4mm}
\vspace*{1mm}
	\begin{minipage}[b]{1.0\linewidth}
		\centering
		\tikzstyle{block} = [draw, fill=none, rectangle, 
		minimum height=2em, minimum width=2em]
		\tikzstyle{sum} = [draw, fill=none, circle, node distance=1cm]
		\tikzstyle{cir} = [draw, fill=none, circle, line width=1mm, minimum width=0.7cm, node distance=1cm]
		\tikzstyle{loss} = [draw, fill=none, color=black, ellipse, line width=0.5mm, minimum width=0.7cm, node distance=1cm]
		\tikzstyle{blueloss} = [draw, fill=none, color=black, ellipse, line width=0.5mm, minimum width=2.5cm, node distance=1cm, color=black]
		\tikzstyle{input} = [coordinate]
		\tikzstyle{output} = [coordinate]
		\tikzstyle{pinstyle} = [pin edge={to-,thin,black}]
		\begin{tikzpicture}[auto, node distance=2cm,>=latex']
		cloud/.style={
			draw=red,
			thick,
			ellipse,
			fill=none,
			minimum height=1em}
		\node [input, name=input] {};
		\node [cir, node distance=1.2cm, right of=input] (Y) {$\rr$};
		\node [block, node distance=1.2cm, right of=Y] (phi_T) {$\proj^{\text{T}}$};
		\node [block, node distance=1.2cm, right of=phi_T] (A_T) {$\frac{1}{L}\A^{\text{T}}$};
		\node [sum, right of=A_T, node distance=1.2cm] (sum) {$+$};
		\node [block, node distance=1.2cm, right of=sum] (relu) {$\includegraphics[width=0.04\linewidth]{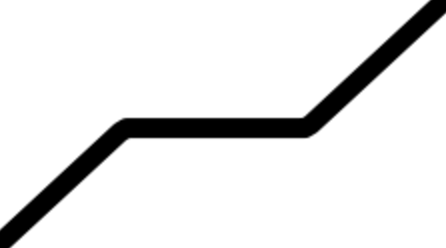}$};

		\node [cir, right of=relu, node distance=1.2cm] (xt) {$\x_{t}$};
		\node [cir, right of=xt, node distance=1.5cm] (xT) {$\x_{T}$};
		
		\node [output, node distance=1.2cm, right of=xT] (output) {};
		
		\node [block, right of=xT, node distance=1.2cm] (A) {$\A$};
		\node [block, right of=A, node distance=1.2cm] (phi) {$\proj$};
		\node [cir, right of=phi, minimum width=0.7cm, node distance=1.2cm] (r_hat) {$\hat \rr$};
		\node [cir, below of=r_hat, minimum width=0.7cm, node distance=0.87cm] (Y1) {$\rr$};
		\node [block, below of=relu, node distance=0.8cm] (f) {$g(\cdot)$};
		
		\node [rectangle, fill=none,  node distance=0.6cm,  below of=r_hat] (middle) {};
		\node [blueloss, right of=middle, node distance=1.8cm] (mse) {loss $\mathcal{L}_{\A}$};

		\draw[thick, line width=2, black, ->]     ($(xt.north east)+(1.5,0.55)$) -- ($(Y.north east)+(11.0,0.55)$);
		\draw[thick, line width=2, black, ->]     ($(Y.north east)+(0,0.55)$) -- ($(xt.north east)+(1.0,0.55)$);

		\node [block, below of=xT, node distance=2cm] (C) {$\C$};
		\node [cir, below of=relu, minimum width=0.7cm, node distance=3.00cm] (one) {$\textbf{1}$};
		\node [block, below of=xt, node distance=3.00cm] (d) {$\dee$};	
		\node [cir, below of=Y1, minimum width=0.7cm, node distance=1.25cm] (u) {$\smlu$};
		\node [sum, below of=C, node distance=0.98cm] (sum_C) {$+$};
		
		\node [cir, below of=u, minimum width=0.7cm, node distance=0.87cm] (u_hat) {$\hat \smlu$};
		
		\node [block, left of=u_hat, node distance=1.53cm] (softmax) {$\text{softmax}$};	
		\node [rectangle, fill=none,  node distance=0.6cm,  below of=u] (middle_class) {};
		\node [blueloss, right of=middle_class, node distance=1.8cm] (class) {loss $\mathcal{L}_{\C,\dee}$};

		
		\draw[thick,dotted]     ($(xT.north east)+(-1.15,0.13)$) rectangle ($(sum.south west)+(-0.3,-1.0)$); 
		\node [rectangle, fill=none,  node distance=1.05cm,  right=20pt,above of=A] (text) {\footnotesize{Decoder}};
		\node [rectangle, fill=none,  node distance=1.05cm,  right=10pt,above of=sum] (encoder) {\footnotesize{Encoder}};
		\draw[thick, line width=2, black, ->]     ($(sum.north east)+(0.8,-3.85)$) -- ($(Y.north east)+(11.0,-3.85)$);
		\node [rectangle, fill=none,  node distance=1.8cm,  right=20pt,below of=C] (text) {\footnotesize{Classifier}};
		
		\draw [->] (Y) -- node [name=m, midway, below] {} (phi_T);
		\draw [->] (phi_T) -- node {} (A_T);
		\draw [->] (A_T) -- node {} (sum);
		\draw [->] (sum) -- node[] {} (relu);
		\draw [->] (relu) -- node[] {} (xt);
		\draw [->] (xt) -- node[name=loop, pos=0.3, above] {} (xT);
		\draw [->] (xT) -- node[] {} (A);
		\draw [->] (A) -- node[] {} (phi);
		\draw [->] (phi) -- node[] {} (r_hat);
		\draw [->] (loop) |- node[] {} (f);
		\draw [->] (f) -| node[pos=1,right] {$$} (sum);

		\node [rectangle, fill=none,  node distance=0.6cm,  above of=relu] (text) {\footnotesize{Repeat $T$ times}};
		
		\draw [black, dashed, thick] (r_hat) --  (mse);
		\draw [black, dashed, thick] (Y1) --  (mse);

		\draw [->] (xT) -- node[] {} (C);
		\draw [->] (C) -- node[] {} (sum_C);
		\draw [->] (one) -- node[] {} (d);
		\draw [->] (d) -- node[] {} (sum_C);
		\draw [->] (sum_C) -- node[] {} (softmax);
		\draw [->] (softmax) -- node[] {} (u_hat);
		
		\draw [black, dashed, thick] (u_hat) --  (class);
		\draw [black, dashed, thick] (u) --  (class);
		
		\draw[thick,dotted]     ($(Y.north east)+(-0.88,0.9)$) rectangle ($(r_hat.south east)+(+3,-1.05)$);
		\node [rectangle, fill=none,  node distance=2.2cm,  left of=Y] (title) {\footnotesize{Unsupervised Block}};
		
		\draw[thick,dotted]     ($(C.north east)+(-3.7,0.2)$) rectangle ($(u_hat.south east)+(+3,-0.70)$);
		\node [rectangle, fill=none,  node distance=2.15cm,  left of=one] (title) {\footnotesize{Supervised Block}};
		
				
		\end{tikzpicture}
	\end{minipage}
	\vspace*{-8mm}
	\caption{RandNet architecture, where $g(\cdot) = (\eye - \frac{1}{L} \A^{\text{T}}\proj^{\text{T}}\proj\A)(\x_t + \frac{s_t - 1}{s_{t+1}} (\x_t - \x_{t-1}))$ and $\rr = \proj \y$. The unsupervised block is a CRsAE architecture (Sec.~\ref{sec:crsae}) when $\proj = \eye$.}
	\label{fig:arc}
	\vspace*{-5mm}
\end{figure*}

\noindent where $J$ is the total number of vectors and $\smlv^j$ is i.i.d. noise. In what follows, we refer to each $\y^j$ as an \emph{example}, each $\x^j$ as a sparse \emph{code}, and assume the entries of $\smlv^j$ have variance $\sigma^2$. The goal of DL is to learn a dictionary $\A$ such that each example $\y^j$ can be approximated as the sparse linear combination of columns from $\A$ using the sparse vector $\x^j$. From an optimization perspective, DL problem aims to solve
\vspace{-2mm}
\begin{equation} \label{eq:altmin1}
\min_{\substack{(\x^j)_{j=1}^J\\ \A}}\ \sum_{j=1}^J \frac{1}{2}\|\y^j - \A \x^j\|^2 + \lambda \|\x^j\|_1\ \text{ s.t. } \|\smla_i\|_2 = 1
\vspace{-2mm}
\end{equation}
\noindent for $i = 1,\cdots, p$, where $\lambda>0$ is a sparsity enforcing parameter that depends on the statistics of the noise $\smlv^j$. The vector $\smla_i$ denotes the $i^{th}$ column of the matrix $\A$, and the norm constraint is to avoid scaling ambiguity. The objective is jointly non-convex in $\A$ and $\{\x^j\}_{j=1}^J$. 
\vspace{-4mm}
\subsection{Classical DL}
\vspace{-1mm}
A classical method to circumvent the non-convexity in Eq.~\ref{eq:altmin1} is the \emph{alternating-minimization} algorithm~\cite{Agarwal2016LearningSU}, which alternates between a \emph{sparse coding} step and a \emph{dictionary update} step. Given an estimate of the dictionary, the sparse coding step estimates each sparse code $\x^j$ by solving
\vspace{-1mm}
\begin{equation} \label{eq:sc}
\min_{\x^j}\ \frac{1}{2}\|\y^j - \A \x^j\|^2 + \lambda \|\x^j\|_1.
\end{equation}
\noindent The dictionary update step uses the newly-estimated codes to update the dictionary as the solution of
\vspace{-2mm}
\begin{equation} \label{eq:dl}
\min_{\A}\ \sum_{j=1}^J \frac{1}{2}\|\y^j - \A \x^j\|^2\ \text{ s.t. } \|\smla_i\|_2 = 1.
\vspace*{-4mm}
\end{equation}
\subsection{Constrained Auto-Encoders}\label{sec:crsae}
A recent line of work~\cite{TolooshamsBahareh2018SCDL,TolooshamsBahareh2019deepresidualAE} has developed an auto-encoder, named CRsAE, to solve Eq.~\ref{eq:altmin1} when $\A$ is a convolutional operator. The encoder in CRsAE imitates the sparse coding step: it uses $\A$, $\A^{\text{T}}$, and the ReLU activation function 
to map the data to a sparse code. The decoder is linear and uses $\A$, constrained to the matrix used in the encoder, to map the output of the encoder to a reconstruction of the data. To learn $\A$, CRsAE minimizes the squared-error reconstruction loss by backpropagation.  
The unsupervised block of Fig.~\ref{fig:arc}, when the matrix $\proj$ is the identity, illustrates the CRsAE architecture. We extend the work in~\cite{TolooshamsBahareh2018SCDL, TolooshamsBahareh2019deepresidualAE} and show, for the first time, that we can train CRsAE to learn a dense dictionary. In addition, through a connection with compressive DL~\cite{Pourkamali-AnarakiFarhad2015Edlv,compKSVD}, 
we show that we can train a modified CRsAE, termed RandNet, to perform DL using randomly-compressed versions of the data $\y^j$.

\vspace*{-3mm}
\subsection{DL from Compressed Data}

Compressive DL is the problem of learning a dictionary from compressed data, where we assume the compressed data are obtained by projecting the original data into a lower dimensional subspace~\cite{Pourkamali-AnarakiFarhad2015Edlv,compKSVD}. Compressive DL~\cite{compKSVD} uses a version of alternating-minimization based on the KSVD algorithm and an $\ell_0$ constraint to enforce sparsity. Using the $\ell_1$-norm, the problem becomes 
\vspace{-2mm}
\begin{equation} \label{eq:altmin2}
\min_{\substack{(\x^j)_{j=1}^J\\ \A}}\ \sum_{j=1}^J \frac{1}{2}\|\rr^j - \proj\A \x^j\|^2 + \lambda \|\x^j\|_1\ \text{ s.t. } \|\smla_i\|_2 = 1,
\vspace{-2mm}
\end{equation}
\noindent where $\rr^j = \proj \y^j$ is the compressed version (random projection) of the example $\y^j$ and $\proj \in \R^{M \times N}$ is a known random measurement matrix such that $M < N$. The work in~\cite{compKSVD} shows that it is indeed possible to learn the dictionary $\A$ when $M < N$ and $\proj$ is a random Gaussian matrix. 
For the rest of the paper, we drop the superscript $j$ to simplify notation.

\vspace{-3mm}
\section{RandNet}
\label{sec:randnets}
\vspace{-2mm}
We introduce a class of NNs, which we call RandNet, for learning representations from compressed data. First, we introduce the architecture in unsupervised and supervised settings. In unsupervised settings, RandNet is a variant of the CRsAE architecture (Sec.~\ref{sec:crsae}) to solve Eq.~\ref{eq:altmin2}. In supervised cases, RandNet uses the output of the encoder as features for the supervised task. Then, we explain why RandNet is memory and computationally efficient, both when the measurement matrix is Gaussian and when it is row sparse. 

\vspace*{-3mm}
\subsection{Unsupervised NNs with Compressed Inputs}\label{sec:unsup}
\vspace{-1mm}
RandNet in unsupervised settings is an auto-encoder. Given $\rr$, $\proj$ and $\A$, the encoder solves the sparse coding step 
in Eq.~\ref{eq:altmin2}. The encoder implements the FISTA algorithm~\cite{beck2009fast}, which generates a sequence $\x_t$, indexed by iteration number $t$, that converges to a sparse code $\x_T$ after $T$ iterations.
Similar to~\cite{TolooshamsBahareh2018SCDL}, we define a state vector $\z_t = \begin{bmatrix}\z_{t}^{(1)}\  \z_{t}^{(2)}\end{bmatrix}^{\text{T}} = \begin{bmatrix}\x_{t}\quad \x_{t-1}\end{bmatrix}^{\text{T}}$. 
Algorithm~\ref{algo:encoder} details the steps of the encoder,  where the two-sided ReLU non-linearity $\eta_{\epsilon}: \R^{p} \to \R^{p}$ is defined element-wise as $(\eta_{\epsilon}(\z))_n = (|z_n|-\epsilon)_+ \textrm{sgn}(z_n)$. The decoder applies $\A$ and then $\proj$ to $\x_T$ to obtain $\cvec_{T+1} = \hat \rr = \proj \A \x_T$. Similar to CRsAE, the parameters of the encoder and decoder are tied. The unsupervised block in Fig.~\ref{fig:arc} shows this forward pass. The loss function associated with the architecture is $\mathcal{L}_{\A}({\rr, \hat \rr)} =  \half \|\rr - \hat \rr\|_2^2$.

\begin{algorithm}
\KwIn{$\rr, \A, \proj, \lambda, L \geq \sigma_\text{max}(\A^{\text{T}}\A)$}
\KwOut{$ \x_T$}
$\z_0 = \mathbf{0},s_0 = 0$\\
\For{$t =1$ to $T$}{
$s_t = \frac{1 + \sqrt{1+4s_{t-1}^2}}{2}$\\
$\w_t = \begin{bmatrix}\left(1 + \frac{s_{t-1}-1}{s_t}\right) \eye_{p}| - \frac{s_{t-1}-1}{s_t} \eye_{p}\end{bmatrix} \z_{t-1}$ \\
$\cvec_t = \w_t + \frac{1}{L}\A^{\text{T}}\proj^{\text{T}}(\rr-\proj \A\w_t)$\\
$\z_t =  \begin{bmatrix}\x_{t}\quad \x_{t-1}\end{bmatrix}^{\text{T}} = \begin{bmatrix}\eta_{\frac{\lambda}{L}}(\cvec_t) \quad \z_{t-1}^{(1)}\end{bmatrix}^{\text{T}}$
}
\caption{Encoder of RandNet.}
\label{algo:encoder}
\end{algorithm}

\vspace{-7mm}
\subsection{Supervised NNs with Compressed Inputs}
In supervised settings, e.g.~classification, we extend RandNet following a similar approach to the architecture in~\cite{Rolfe2013DiscriminativeRS}. Specifically, we show how a network should be designed for a classification task to be trained on compressed data. For $K$-class classification, we define a $K$-dimensional label vector $\smlu \in \{0,1\}^{K}$ such that for each class $c$
\vspace*{-2mm}
\begin{equation}\label{eq:class_label}
u_k = \begin{cases}
    1, & \text{if $k = c$};\\
    0, & \text{otherwise}.
  \end{cases}
\vspace{-2.0mm}
\end{equation}
\noindent The goal is to learn a mapping from the sparse representation $\x_T$ of the data produced at the output of the encoder (see Sec.~\ref{sec:unsup}) to an estimated vector of probabilities $\hat \smlu = \frac{e^{\smlq}}{\sum_i e^{\smlq_i}}$, where $\smlq = \C \x_T + \dee$. The supervised block in Fig.~\ref{fig:arc} shows the forward pass of the resulting architecture. We define the categorical cross-entropy loss
\vspace*{-2mm}
\begin{equation}\label{eq:loss_supervised}
\mathcal{L}_{\C,\dee}{(\x_T, \smlu, \C, \dee)} =   - \smlu^{\text{T}} \log{\left(\frac{e^{\C \x_T + \dee}}{\sum_i e^{(\C \x_T + \dee)_i}}\right)}.
\vspace{-3.0mm}
\end{equation}

\subsection{RandNet Backpropagation Algorithm}

\noindent In the unsupervised case, we minimize $\mathcal{L}_{\A}({\rr, \hat \rr)}$ with the constraint $\|\smla_i\|_2 = 1$ to avoid scaling ambiguity between the dictionary and sparse codes.
For the supervised case, we proceed in two stages to optimize the parameters of interest. First, we train the unsupervised network to learn $\x_T$ that approximates the compressed data when multiplied by $\proj \A$. Then, given $\x_T$, we learn the matrix $\C$ and the bias $\dee$ that minimize the loss $\mathcal{L}_{\C,\dee}{(\x_T, \smlu, \C, \dee)}$.

Algorithm~\ref{algo:bprop} is the backpropagation algorithm for computing the gradient, denoted $\delta\cdot$, of the loss functions with respect to the parameters of interest. The vectors $\smla = [\smla_1^{\text{T}}\ \smla_2^{\text{T}}\ \cdots \smla_p^{\text{T}}]^{\text{T}}$ and $\tilde \cvec = [\cvec_1^{\text{T}}\cvec_2^{\text{T}}\ \cdots \cvec_K^{\text{T}}]^{\text{T}}$ are the vectors of stacked columns from $\A$ and $\C$, respectively. 
In practice, we compute the gradients through PyTorch's autograd. We omit the derivation.

\vspace{-3mm}
\begin{algorithm}
\KwIn{$\rr, \smlu, \lambda,L,\A, \proj, \C, \dee$, Variables $s_t$, $\w_t$, $\cvec_t$, $\z_T$, $\x_T$ from RandNet encoder, and $\hat{\rr}, \cvec_{T+1}$.}
\KwOut{$\delta \smla, \delta \tilde \cvec, \delta \dee$}
$\delta \hat \rr = \hat \rr - \rr, (\delta \hat \smlu)_k = -\frac{u_k}{\hat u_k}, \delta \smla = \mathbf{0}_{Np}$\\
$\delta \smlq = \frac{\partial \hat \smlu}{\partial \smlq} \delta \hat \smlu $, $\delta \dee = \frac{\partial \smlq}{\partial \dee} \delta \smlq$, $\delta \tilde \cvec = \frac{\partial \smlq}{\partial \tilde \cvec} \delta \smlq$\\
$\delta \cvec_{T+1} = \frac{\partial \hat \rr}{\partial \cvec_{T+1}} \delta \hat \rr $\\
$\delta \smla = \delta \smla +  \frac{\partial \cvec_{T+1}}{\partial \smla} \delta \cvec_{T+1}$\\
$\delta \z_{T} = \frac{\partial \cvec_{T+1}}{\partial \z_{T}} \delta \cvec_{T+1}$\\
\For{$t =T$ to $1$}{
$\delta \cvec_t = \frac{\partial \z_t}{\partial \cvec_t} \delta \z_t $\\
$\delta \smla = \delta \smla +  \frac{\partial \cvec_t}{\partial \smla} \delta \cvec_t$\\
$\delta \z_{t-1} = \frac{\partial \cvec_t}{\partial \z_{t-1}} \delta \cvec_t$
}
\caption{RandNet backprop for $\A$, $\C$, and $\dee$.}
\label{algo:bprop}
\end{algorithm}

\setlength{\tabcolsep}{3pt}
\begin{table} 
\vspace*{-4mm}
  \centering
  \begin{tabular}{lllll}
     &   & (a) Gaussian & (b) Sparse &(c) Identity\\ \midrule
Memory Storage &&  {\boldmath$O(\beta N)$} &  $\qquad \cdot$  & $O(N)$\\ \midrule
Memory Access && $\qquad \cdot$ &  {\boldmath$O(\gamma N)$}  & $O(N)$\\ \midrule

    \multirow{2}{*}{Matrix Operation}  & \multicolumn{1}{l}{$\proj^{\text{T}}$} & \multicolumn{1}{l}{ $O(MN)$}  & \multicolumn{1}{l}{ $O(\gamma N)$} & $\quad \cdot$ \rule{0pt}{2ex}\\ \cline{2-5}\rule{0pt}{2.5ex}
                                 & \multicolumn{1}{l}{$\A^{\text{T}}$} & \multicolumn{1}{l}{ $O(pN)$}  & \multicolumn{1}{l}{ {\boldmath$O(\gamma pN)$}} & \multicolumn{1}{l}{$O(pN)$}  \\ \bottomrule
  \end{tabular}
\caption{Memory and computational efficiency of RandNet when using (a) Gaussian measurements, (b) sparse random measurements to compress images, and (c) no compression.}
\vspace*{-4mm}
\label{tab:comp}
\end{table}

\vspace{-3mm}
\noindent \textbf{Benefits of random projections}: One advantage of RandNet compared to classical NNs is that a larger amount of data can fit on GPU memory, because the network is trained with the compressed data. Let $\beta \overset{\Delta}{=} \frac{M}{N} < 1$ denote the measurement ratio~\cite{compKSVD}. The memory storage cost of RandNet is $O(\beta N)$, as opposed to $O(N)$ for a network without compression. For Gaussian $\proj$, similar to~\cite{compKSVD}, RandNet takes advantage of this reduction in memory storage cost by storing the compressed data instead of the original data. The benefits of RandNet are even more pronounced when $\proj$ is row sparse. In this case, we assume each row of $\proj$ is $s$-sparse ($s \geq 1$) with non-zero entries chosen uniformly at random and set to $\{-1, +1\}$ with equal probability. Let $\gamma \overset{\Delta}{=}  \beta s < 1$ denote the compression factor~\cite{pmlr-v32-anaraki14,Pourkamali-AnarakiFarhad2015Edlv}. The cost for the GPU of accessing the data during training is $O(\gamma N)$, compared to the cost $O(N)$ of accessing the data in its original dimension~\cite{pmlr-v32-anaraki14}. Therefore, in the sparse case and for small enough $\gamma$, the cost of accessing the data is even lower than storing and accessing it in compressed form. In addition to storage/access benefits, row-sparse $\proj$ make RandNet efficient computationally.  In RandNet, the cost of applying $\A^{\text{T}}$ is $O(\gamma pN)$ compared to $O(pN)$ without compression. This is because $\A^{\text{T}}$ operates on the sparse vector produced by the column-sparse operator $\proj^\text{T}$ (Algorithm~\ref{algo:encoder}, line $5$). Table~\ref{tab:comp} summarizes these benefits.


\noindent \textbf{Shared projection operators}: In practice, similar to~\cite{Pourkamali-AnarakiFarhad2015Edlv}, we divide the data into $B$ blocks. Examples with indices $j = \left\{\frac{J}{B} (b-1) + 1, \cdots, b\frac{J}{B}\right\}$ belong to block $b$ and share the measurement matrix $\proj_b$, $b = 1\cdots, B$. 
We denote an example from block $b$ as $\y^b$ and its projection $\rr^b$. We note that we do not need to store the $\proj_b$ matrices. Instead, we store the fixed random seed associated with each matrix. 
For the rest of the paper, we drop the notation $b$ to simplify the notation. 

\vspace{-4mm}
\section{Experiments}
\label{sec:expts}
\vspace{-2mm}
We train RandNet on a simulated dataset and on MNIST. In the simulated case, we train a) a CRsAE architecture to learn the dictionary underlying data simulated according to Eq.~\ref{eq:gen}, and b) an unsupervised RandNet architecture to learn the same dictionary from random projections of the data when $\proj$ is Gaussian and also when it is row sparse. As a benchmark, we implement the CK-SVD algorithm~\cite{Pourkamali-AnarakiFarhad2015Edlv}.
We use MNIST to demonstrate the use of RandNet in a classification task with randomly-projected data (digits) as inputs. 

\vspace{-3mm}
\subsection{Datasets}
\vspace{-2mm}
\subsubsection{Simulation}
\vspace{-2mm}
\noindent We simulated a dataset of  $J=4{,}250$ examples $\y^j \in \R^{500}$ from the generative model of Eq.~\ref{eq:gen} with $\smlv^j = 0$. The elements of $\A \in \R^{500 \times 20}$ are i.i.d.~and drawn according to a $\mathcal{N}(0,\frac{1}{500})$ distribution, followed by the normalization of each column to have unit length. Each sparse code $\x^j$ is $3$-sparse with i.i.d.~nonzero entries following a Uniform distribution on the interval $[-5,-4] \cup [4,5]$. We generated $B = 40$ measurement matrices $\{\proj_b \in \R^{M \times 100}\}_{b=1}^B$. We used either Gaussian  matrices or sparse ones with $1$-sparse rows $(s = 1)$. To study the effects of compression, we considered $M=250, 150, 50$, i.e.~$\beta = 0.5, 0.3, 0.1$, respectively. The corresponding compression factors are $\gamma = 0.5, 0.3, 0.1$, respectively. 
For Gaussian $\proj$, RandNet reduces the cost of storage by a factor of $\beta$. For row-sparse $\proj$, both the memory access cost and operation cost of $\A^{\text{T}}$ are reduced by a factor of $\gamma$.

\vspace{-2mm}
\subsubsection{MNIST}
We considered the MNIST dataset comprising  $70{,}000$ $28 \times 28$ grayscale handwritten digits, split into a training set with $60{,}000$ images and a test set with $10{,}000$ images. We vectorized the images, resulting in $\{\y^j \in \R^{784}\}_{j=1}^J$, $J=70{,}000$. We generated $B = 1{,}000$ random measurement matrices $\{\proj_b \in \R^{M \times 784}\}_{b=1}^B$, for $M=392$ ($\beta = \gamma = 0.5$). 
The goal is to learn a dictionary $\A \in \R^{784 \times 784}$ that yields a $784$-dimensional sparse representation of the images that is useful for classifying each image into one of $K = 10$ categories. Hence, the classification matrix $\C \in \R^{10 \times 784}$.
\vspace{-3mm}
\subsection{Training}
We implemented RandNet on PyTorch and trained it on a GPU with backpropagation (Tesla V100-SXM2) by mini-batch gradient descent. We used the ADAM optimizer.
\vspace{-4mm}
\subsubsection{Simulation}
\vspace{-1mm}
We divided the data into a training set of size $4{,}000$ and a test set of size $250$. The number of trainable parameters is $500 \times 20 = 10{,}000$. We set the number of FISTA iterations to $T = 400$. This is crucial in producing sparse codes, which facilitates DL~\cite{Agarwal2016LearningSU,TolooshamsBahareh2019deepresidualAE}. We set $\lambda = \sigma \sqrt{2 \log{p}}$~\cite{Chen1998AtomicDB} and tuned $\sigma$ by grid search over the interval $[0.01, 0.2]$. We set $L = 5$, $12$, and $2$ for CRsAE, Gaussian, and sparse RandNet, respectively. We picked each of these values so that it is greater than the maximum eigenvalue of $\A^{\text{T}}\proj^{\text{T}}\proj \A$~\cite{beck2009fast}, where $\proj$ depends on the architecture. We used a batch size of $64$, a learning rate of $0.001$, and trained for $10$ epochs.
\vspace{-5mm}
\subsubsection{MNIST}
\vspace{-2mm}
We first trained the RandNet auto-encoder for $20$ epochs, and then the classifier for another $20$ epochs. We hypothesized that the sparse code $\x_T$ obtained at the output of the encoder would be a useful representation for classification. The number of trainable parameters is $784 \times 784 = 614{,}656$ for the auto-encoder, and $10 \times 784 + 10$ for the classifier, a total of $622{,}506$ parameters. We set the number of FISTA iterations to $T=60$. We tuned $\lambda$ by grid search and used $\lambda = 2.2$ and $\lambda=2$ for the Gaussian and sparse cases, respectively. We used $L = 50$ and estimated it to be greater than the maximum eigenvalue of $\A^{\text{T}}\proj^{\text{T}}\proj \A$ when $\A$ is a random Gaussian matrix. We used a batch size of $16$ and a learning rate of $0.005$. We initialized the estimated dictionary $\hat{\A}$ with i.i.d.~$\mathcal{N}(0,\frac{1}{784})$ entries and normalized its columns to have unit length. We initialized $\hat{\C}$ and $\hat{\dee}$ with i.i.d.~Uniform entries in the interval $[\frac{-1}{\sqrt{784}},\frac{1}{\sqrt{784}}]$.

\vspace*{-2mm}
\begin{figure}[htb]
\begin{minipage}[b]{1.0\linewidth}
  \centering
  \centerline{\includegraphics[width=8.5cm]{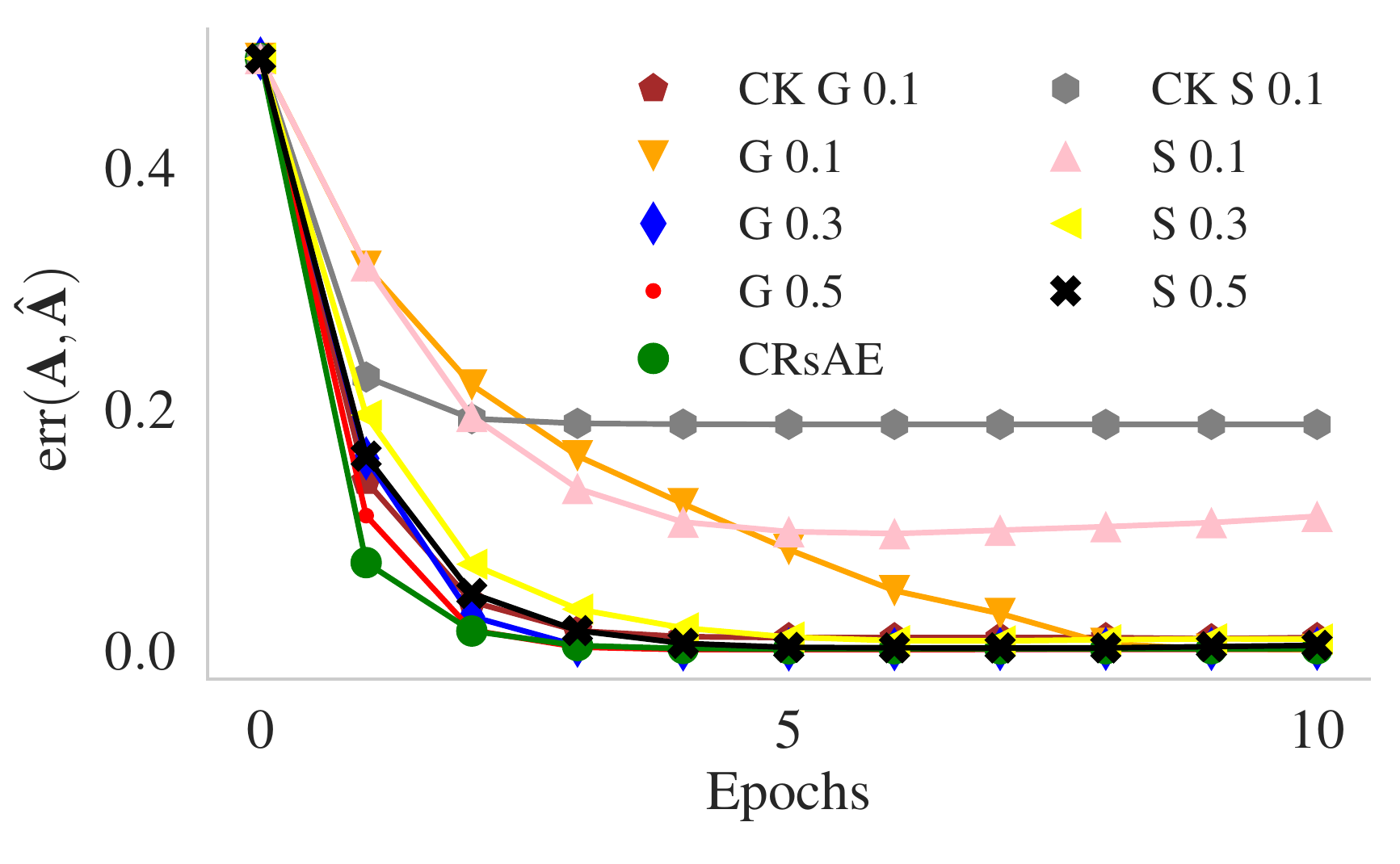}}
\end{minipage}
\vspace*{-10mm}
\caption{Error $\text{err}(\A,\hat \A)$ as a function of epoch for CRsAE (green $\circ$) and RandNet. ``G 0.1'' stands for RandNet with Gaussian $\proj$ and $\beta = 0.1$. ``S 0.5'' stands for RandNet with sparse $\proj$ and $\beta = 0.5$. ``CK'' stands for CK-SVD.}
\label{fig:sim_err}
\vspace*{-3mm}
\end{figure}

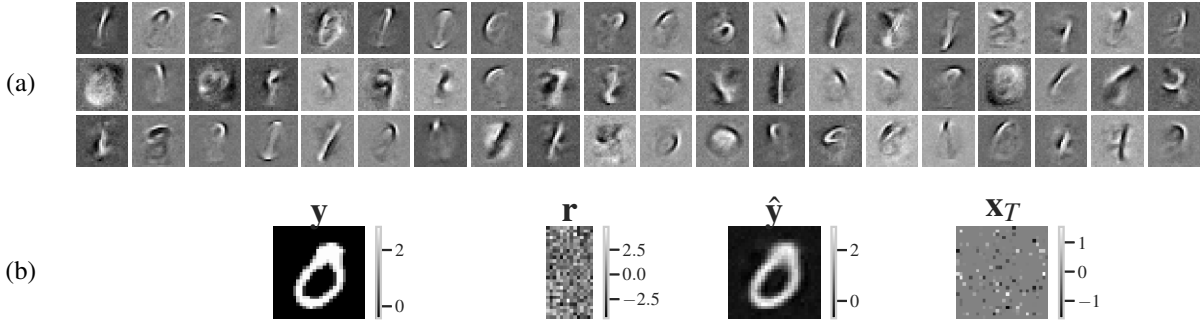
\begin{figure*}[h]
	\vspace*{-4mm}
	\begin{minipage}[b]{1.0\linewidth}
		\centering
		\tikzstyle{block} = [draw, fill=none, rectangle, 
		minimum height=2em, minimum width=2em]
		\tikzstyle{sum} = [draw, fill=none, circle, node distance=1cm]
		\tikzstyle{cir} = [draw, fill=none, circle, line width=1mm, minimum width=0.7cm, node distance=1cm]
		\tikzstyle{loss} = [draw, fill=none, color=black, ellipse, line width=0.5mm, minimum width=0.7cm, node distance=1cm]
		\tikzstyle{blueloss} = [draw, fill=none, color=black, ellipse, line width=0.5mm, minimum width=2.5cm, node distance=1cm, color=black]
		\tikzstyle{input} = [coordinate]
		\tikzstyle{output} = [coordinate]
		\tikzstyle{pinstyle} = [pin edge={to-,thin,black}]
		\begin{tikzpicture}[auto, node distance=2cm,>=latex']
		cloud/.style={
			draw=red,
			thick,
			ellipse,
			fill=none,
			minimum height=1em}
		\node [input, name=input] {};
		
		\node [rectangle, fill=none, node distance=3cm, right of=input] (C) {$\includegraphics[width=0.85\linewidth]{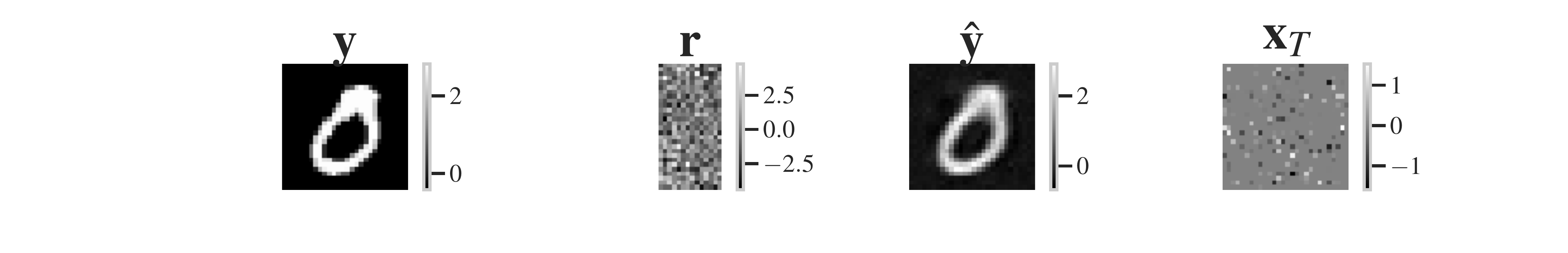}$};
		\node [rectangle, fill=none, node distance=2.5cm, above of=C] (A) {$\includegraphics[width=0.885\linewidth]{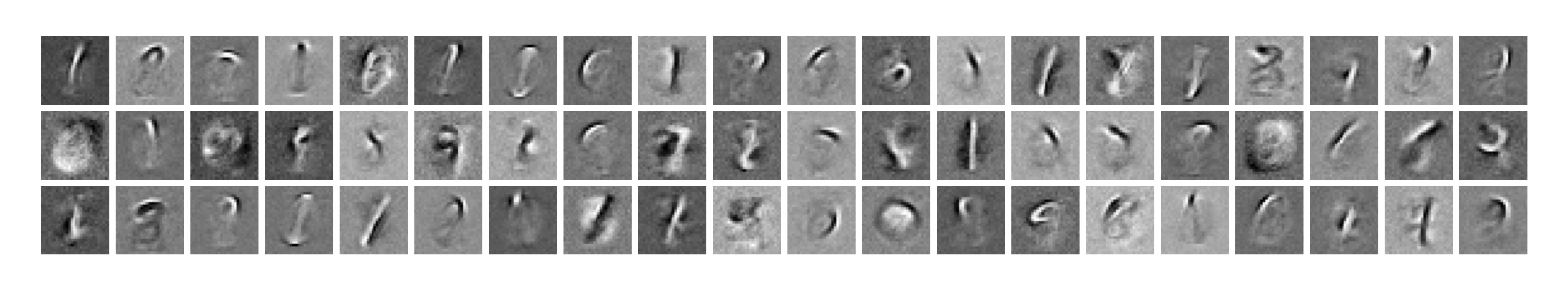}$};
				
		\node [rectangle, fill=none,  node distance=8.2cm,  left of=A] (text) {(a)};
		\node [rectangle, fill=none,  node distance=8.2cm,  left of=C] (text) {(b)};

		\end{tikzpicture}
	\end{minipage}
	\vspace*{-15mm}
	\caption{(a) The $60$ columns of the learned dictionary $\hat{\A}$ used most frequently to reconstruct the compressed MNIST digits  when $\proj$ is Gaussian and $\beta = 0.5$ . (b) RandNet applied to an instance of digit $0$. Left: digit in its original and compressed dimensions. Right: reconstruction and sparse representation of the digit. Note that $\x_T$ is highly sparse (most entries close to zero).}
	\label{fig:mnist_A}
	\vspace*{-5mm}
\end{figure*}

\vspace*{-6mm}
\subsection{Results}
\vspace*{-2mm}
\subsubsection{Simulation}
We use the following measure~\cite{Agarwal2016LearningSU} to quantify the distance between the learned dictionary and that used to simulate the data
\vspace*{-3mm}
\begin{equation}\label{eq:distance_error}
\text{err}(\A,\hat \A) = \max_i \left(\sqrt{1 - \frac{\ip{\smla_i}{\hat{\smla}_i}^2}{\|{\smla}_i\|_2^2\|\hat{\smla}_i\|_2^2}}\right).
\vspace*{-1mm}
\end{equation}
The lower this measure, which ranges from $0$ to $1$, the closer the estimated dictionary $\hat{\A}$ to the true dictionary $\A$. We initialized $\hat{\A}$ by randomly perturbing $\A$ so that $\text{err}(\A,\hat \A) \approx 0.5$~\cite{Agarwal2016LearningSU}. Fig.~\ref{fig:sim_err} shows $\text{err}(\A,\hat \A)$ as a function of epochs for CRsAE (green $\circ$), as well as RandNet with Gaussian $\proj$ (G) and row-sparse $\proj$ (S), for $\beta = 0.1$, $0.3$, and $0.5$. The figure highlights the ability of CRsAE to learn the underlying dictionary when the dictionary is dense, and more importantly, the ability of RandNet to learn the dictionary from randomly-projected data that live in a subspace whose dimension is a factor $\beta$ lower than the original dimension. The figure shows that for the case of $\beta = 0.1, 0.3, 0.5$ and Gaussian $\proj$, RandNet successfully learns the dictionary. When $\proj$ is row sparse, we are able to learn the dictionary using RandNet for $\beta = 0.3, 0.5$, and perform better than CK-SVD (CK) for $\beta = 0.1$.

\vspace*{-2mm}
\subsubsection{MNIST}
\noindent \textbf{Unsupervised} Fig.~\ref{fig:mnist_A}(a) shows the $60$ columns of the learned matrix $\hat{\A}$ that are used the most to approximate $\rr$ using $\x_T$. RandNet has learned features similar to the digits in the original data space even though it was trained using randomly-projected data that visually look like noise. Fig.~\ref{fig:mnist_A}(b) is an example of how RandNet processes its inputs.

\noindent \textbf{Supervised} We benchmarked RandNet against CK-SVD, the discriminative recurrent sparse auto-encoder (DrSAE)~\cite{Rolfe2013DiscriminativeRS} architecture, and the supervised DL (SDL) algorithm from~\cite{marialSDL}. 
For CK-SVD, we set the sparsity level to $15$ and learned $\A_c \in \R^{78}$ independently for each class $c$ over 180 iterations. We assign a randomly-projected image to the class with minimum reconstruction error $\|\rr - \proj\A_c \x\|^2$. DrSAE implements a variant of the ISTA algorithm~\cite{beck2009fast} 
and aims to reconstruct the input data, as well as achieve high classification accuracy. SDL jointly solves a DL problem and a classification problem. Table~\ref{tab:mnist_err} shows that RandNet with $\beta = 0.5$ achieves test error rates of $1.56\%$ and $3.16\%$ using Gaussian and row-sparse $\proj$. The corresponding rates for CK-SVD are $3.72\%$ and $5.20\%$, respectively. 
DrSAE and SDL reach $1.08$ and $1.05$ error rates, respectively. For Gaussian $\proj$, the fact that RandNet achieves such a low error using a dataset half the size of the original one is both surprising and impressive. We can interpret the higher error rate in the sparse case as the cost of computational efficiency. The accuracy for more efficient networks (lower $\beta$) was lower and hence not reported.  Below, we explain the higher error rate in the sparse case from the perspective of compressed sensing. 


\vspace*{-2mm}
\begin{table} [htb]
  \centering
   \begin{tabular}{ccccc}
        &   \textbf{RandNet} & CK-SVD & DrSAE & SDL\\ \midrule
    \multirow{2}{*}{Error Rate [$\%$] }  & \multicolumn{1}{c}{$\textbf{(G)}$ $\textbf{1.56}$} & \multicolumn{1}{c}{(G) $3.72$} & \multirow{2}{*}{$1.08$} & \multirow{2}{*}{$1.05$}\\ \cline{2-3}
                                 & \multicolumn{1}{c}{$\textbf{(S)}$ $\textbf{3.16}$} & \multicolumn{1}{c}{(S) $5.20$} &  \\ \bottomrule
  \end{tabular}
\caption{MNIST classification error [$\%$] on test dataset. (G) stands for Gaussian and (S) for row-sparse projection.}
\vspace*{-4mm}
\label{tab:mnist_err}
\end{table}

\noindent \textbf{Why does RandNet work?} 
Suppose an oracle handed us a dictionary $\A$ such that each MNIST image admits a sparse representations in the dictionary. The theory of compressed sensing would then guarantee that the sparse representation can be estimated from very few random projections of the original MNIST images~\cite{candes2008restricted}. The close-to-state-of-the-art performance of RandNet on MNIST is evidence that there indeed exists a dictionary in which MNIST images have sparse representations and, more importantly, that the sparse representations in this dictionary are useful for classification. In the absence of such dictionary, classification of MNIST images from random compressed measurements would likely fail.
\begin{figure}[htb]
\begin{minipage}[b]{1.0\linewidth}
  \centering
  \centerline{\includegraphics[width=8.0cm]{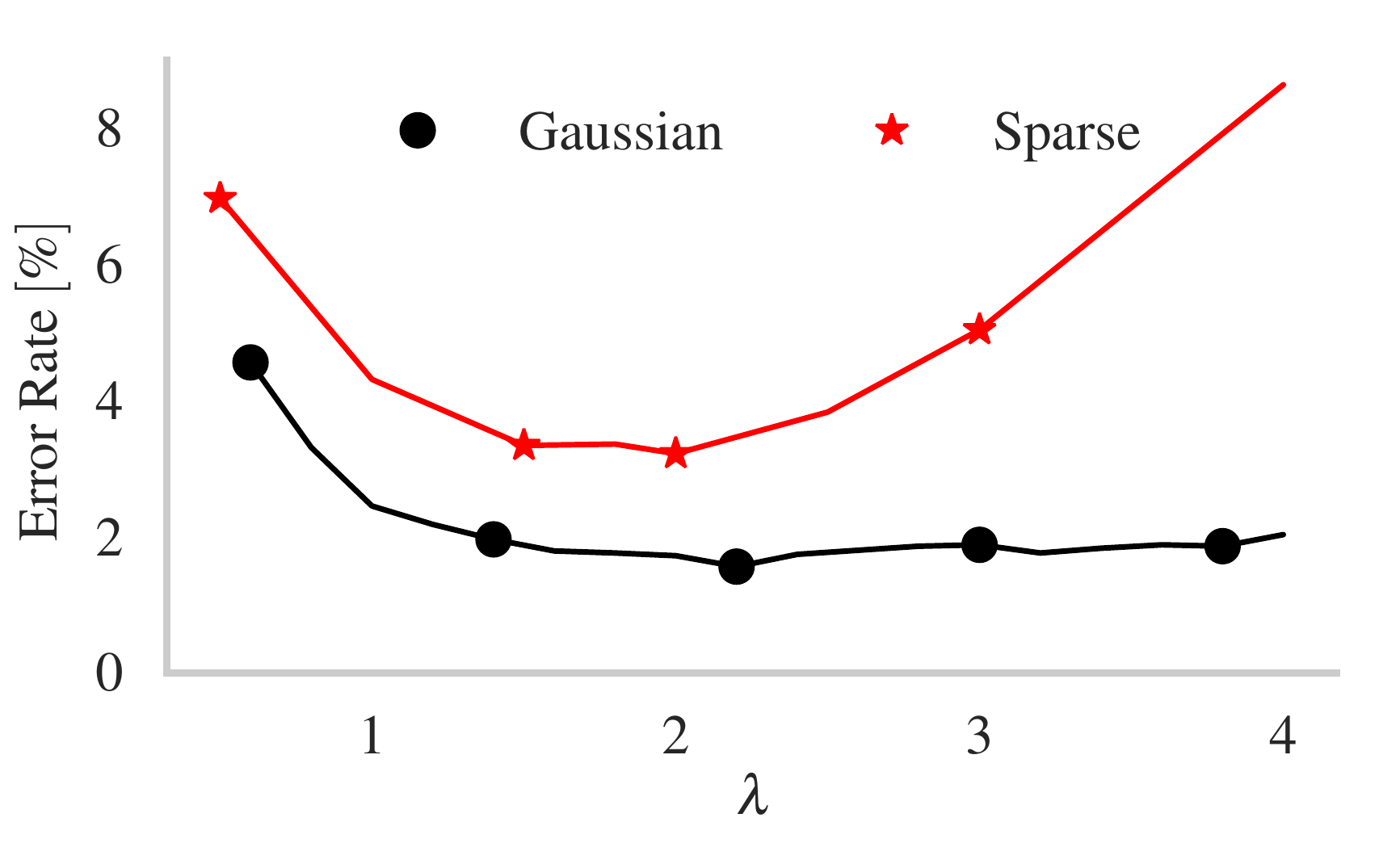}}
\end{minipage}
\vspace*{-10mm}
\caption{MNIST classification error rate $[\%]$ as a function of $\lambda$, which controls the sparsity level of the sparse code $\x_T$ used for classification. RandNet achieved its lowest test error at $\lambda = 2.2$ for Gaussian and $\lambda = 2$ for sparse measurements.}
\label{fig:mnist_lambda}
\vspace*{-5mm}
\end{figure}
To further highlight this intuition, we analyzed the impact of sparsity of the representation $\x_T$ in classification accuracy by sweeping the value of the regularization parameter from $0.5$ to $4$ (the larger this value, the sparser the representation) when $\beta = 0.5$. We found that the classification error reaches its minimum at $\lambda = 2.2$ in the Gaussian case (black $\circ$) and $\lambda=2$ in the sparse case (red $\star$), indicating that there is an optimal amount of sparsity in the learned dictionary that gives good reconstruction and classification accuracies. Fig.~\ref{fig:mnist_lambda} shows such results. 
The compressed-sensing perspective also helps to explain the higher error rate obtained with sparse measurement matrices. For measurement matrices of the same dimension, the RIP constants of sparse matrices are worse than those of dense ones~\cite{BaDemba2018DSrt}. This suggests that, for a sparse measurement matrix, one would need more measurements to achieve the same error rate as for a Gaussian measurement matrix. We hypothesize that decreasing the sparsity of the rows of the measurement matrix will improve classification accuracy. We will explore this in future work.

\vspace*{-3mm}
\section{Conclusion}
\vspace*{-2.5mm}
\label{sec:conc}
We proposed a general framework to train NNs from compressed measurements. Specifically, we introduced RandNet, a class of networks that, in the unsupervised setting, performs dictionary learning from random projections of the original data. In the supervised setting, we highlighted the ability of RandNet in the classification of MNIST when the network is trained using random projections of the images that live in a subspace of smaller dimension compared to the original dimension. RandNet reached an error of $1.56\%$ when the measurement matrix was Gaussian and a $3.16\%$ error rate when using row-sparse measurements, the case where RandNet yields the most significant benefits in terms of memory access and computational efficiency. Overall, RandNet achieved a minimal loss in accuracy considering the increased efficiency in terms of computation and memory.

\vspace*{-3mm}
\section{Acknowledgments}
\vspace*{-3mm}
\noindent This work is partially supported by the Quantitative Biology Initiative at Harvard University.

\vspace*{-3mm}

\bibliographystyle{IEEEbib}
\bibliography{random-nets-arxiv}
\vspace*{-3mm}

\end{document}